\documentclass[sn-mathphys,Numbered]{sn-jnl}

\usepackage{CJKutf8}
\usepackage{graphicx}%
\usepackage{multirow}%
\usepackage{amsmath,amssymb,amsfonts}%
\usepackage{amsthm}%
\usepackage{mathrsfs}%
\usepackage[title]{appendix}%
\usepackage{xcolor}%
\usepackage{textcomp}%
\usepackage{manyfoot}%
\usepackage{booktabs}%
\usepackage{algorithm}%
\usepackage{algorithmicx}%
\usepackage{algpseudocode}%
\usepackage{listings}%

\theoremstyle{thmstyleone}%
\theoremstyle{thmstyletwo}%

\theoremstyle{thmstylethree}%

\begin{document}
\begin{CJK*}{UTF8}{gbsn}
\title[Article Title]{Chinese Financial Text Emotion Mining: GCGTS - A Character Relationship-based Approach for Simultaneous Aspect-Opinion Pair Extraction}
\author{\fnm{Qi} \sur{Chen*}}

\author{\fnm{DeXi} \sur{Liu}}
\equalcont{These authors contributed equally to this work.}


\affil{\orgname{Jiangxi University of Finance and Economics}, \orgaddress{{China}}}

\author{\fnm{} \sur{}}


\abstract{Aspect-Opinion Pair Extraction (AOPE) from Chinese financial texts is a specialized task in fine-grained text sentiment analysis. The main objective is to extract aspect terms and opinion terms simultaneously from a diverse range of financial texts. Previous studies have mainly focused on developing grid annotation schemes within grid-based models to facilitate this extraction process. However, these methods often rely on character-level (token-level) feature encoding, which may overlook the logical relationships between Chinese characters within words. To address this limitation, we propose a novel method called Graph-based Character-level Grid Tagging Scheme (GCGTS). The GCGTS method explicitly incorporates syntactic structure using Graph Convolutional Networks (GCN) and unifies the encoding of characters within the same syntactic semantic unit (Chinese word level). Additionally, we introduce an image convolutional structure into the grid model to better capture the local relationships between characters within evaluation units. This innovative structure reduces the excessive reliance on pre-trained language models and emphasizes the modeling of structure and local relationships, thereby improving the performance of the model on Chinese financial texts. Through comparative experiments with advanced models such as Synchronous Double-channel Recurrent Network (SDRN) and Grid Tagging Scheme (GTS), the proposed GCGTS model demonstrates significant improvements in performance.

}




\maketitle

\section{Introduction}\label{sec1}

Financial texts include not only financial information disclosed by listed companies themselves and other non-financial information, but also various external environmental information that affects the company, as well as interpretations of companies and external environments by major institutions and experts. The information in financial commentary texts can help interpret the operating conditions and environment of listed companies, and the mining of financial commentary texts has received increasing attention. The Aspect-Opinion tuple is the core information of the evaluation text. For example, in the sentence ``原材料 价格 上涨 。''(The price of raw materials has risen), ``原材料 价格''(the price of raw materials) is the aspect term and ``上涨''(rise) is the opinion term, ``$\langle \text{原材料 价格，上涨} \rangle$''($\langle \text{Raw material prices, rise} \rangle$) is the Aspect-Opinion tuple, and the space is the word separator. Aspect-Opinion tuples can more significantly and concisely represent the commentator's point of view, so obtaining high-quality financial information by extracting Aspect-Opinion tuples has gradually become an important task.
\begin{figure}[h]%
\centering
\includegraphics[width=0.5\textwidth]{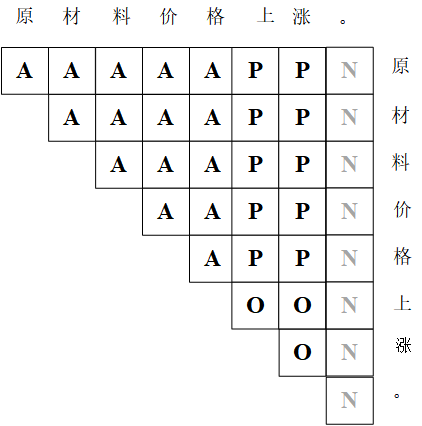}
\caption{An example of using GTS for the AOPE task in Chinese financial text.}\label{fig1}
\end{figure}

Aspect-Opinion Pair Extraction (AOPE), also known as aspect-level sentiment analysis, is an important task in Aspect-Based Sentiment Analysis (ABSA). Currently, there has been rapid progress in research on extracting Aspect-Opinion Pair from product reviews, and the mainstream method to extract sentiment Aspect-Opinion Pair is the grid model \cite{bib1},\cite{bib2},\cite{bib3},\cite{bib4}. The grid model can enumerate the possible aspect term (financial entities and their characteristics), opinion term (the sentiment words for the aspect term), and their relationships in the text. It adopts an end-to-end approach to compensate for the error propagation caused by separating the extraction and pairing of aspect term and opinion term. Taking the grid model \cite{bib1} as an example, the labeling form is shown in Figure \ref{fig1}, where ``A'' represents the aspect term, ``O'' represents the opinion term, ``P'' represents the Aspect-Opinion Pair, and ``N'' represents the non-sentiment element. For convenience in the following discussion, Aspect terms, Opinion terms, and Aspect-Opinion pairs will be collectively referred to as ``evaluation elements''.

However, in Chinese financial text Aspect-Opinion Pair Extraction, grid models have the following shortcomings:

(1) The syntax dependency relationship is not fully utilized in Chinese financial text analysis, especially in identifying the boundaries of evaluation elements. In English text, using syntax analysis to extract evaluation aspects is usually easier because there are clear boundaries between words. However, for Chinese text, there are no obvious separators between Chinese characters, making it difficult to determine the boundaries of evaluation aspects. Previous aspect-level sentiment analysis models have rarely considered this issue. Current research \cite{bib1},\cite{bib2},\cite{bib3},\cite{bib4},\cite{bib5}introduces the use of syntax dependency relationships in modeling the dependencies between words when representing their meanings, but these studies are mostly focused on English datasets such as res14 and res15, and rarely consider using syntax dependency relationships to assist in identifying the syntactic position of Chinese tokens in the sentence to solve the difficult problem of identifying the boundaries of Chinese evaluation elements. For English text, using syntax dependency relationships to extract evaluation aspects is usually easier because there are clear boundaries between words. However, for Chinese text, there are no obvious separators between Chinese characters, making it difficult to determine the boundaries of evaluation aspects. Past aspect-level sentiment analysis models have rarely considered this issue. 

For example, in the sentence ``碘克沙醇 作为 X 光 对比剂 中的 高端 产品 在 医院 市场 中 增长 较快'' (Iodixanol, as a high-end product in X-ray contrast agents, has grown rapidly in the hospital market), when translated into English, the model only needs to identify "iodixanol'' as the evaluation object. However, when ``碘克沙醇'' (iodixanol) is used as a Chinese evaluation object in the input model, it will be split into four tokens with poor relevance: ``碘'' (iodine), ``克'' (gram), ``沙'' (sand), and ``醇'' (alcohol), which makes it difficult for the model to identify the complete evaluation object in the sentence.

(2) Insufficient utilization of Chinese character relationships within evaluation elements. In Chinese financial text, there is a strong correlation between Chinese characters within evaluation elements. The current grid model uses words as the input granularity, and with the help of pre-trained language models, it can learn the semantic information of words well. However, if the input granularity is simply changed from English words to Chinese characters, the grid model may ignore the relationships between Chinese characters within evaluation elements, which ultimately leads to the model's inability to fully learn the boundaries and semantics of evaluation elements.

As shown in Figure \ref{fig1}, the example sentence "原材料价格上涨." (The price of raw materials has risen.), the term "上涨" (rise) is segmented into two characters, "上" (up) and "涨" (rise). They are paired separately with "原材料价格" (the price of raw materials). Due to the lack of proximity information between Chinese characters, the grid model often recognizes only a part of the annotated information. For example, in the case of "上涨" (rise), most of the meaning is concentrated in "涨" (rise), while "上" (up) is often ignored.
Conversely, it is also possible to have the opposite situation where "价格" (price) in "原材料价格" (the price of raw materials) matches better with "上涨" (rise), leading to the neglect of "原材料" (raw materials).
This highlights the importance of proximity information between Chinese characters in Chinese financial text analysis. To address this issue, our research aims to propose an innovative model that can better capture the local relationships between Chinese characters within evaluation elements.

In summary, the existing grid models for Aspect-Opinion Pair Extraction in Chinese financial texts face challenges in effectively utilizing syntax dependency relationships and capturing Chinese character relationships within evaluation elements. These limitations hinder accurate and comprehensive extraction of aspect-opinion pairs from Chinese financial texts. Aiming at the characteristics of Chinese financial commentary texts and the shortcomings of existing grid models, this paper proposes a Chinese financial text aspect opinion pair extraction model, GCGTS, which gives the following contributions:

(1) By employing LAGCN to learn syntactic dependency relationships and embedding the syntactic adjacency matrix into the grid model, the model explicitly utilizes syntax structure, leading to improved accuracy and robustness in recognizing entity boundaries and identifying aspects.

(2) Introducing a flat image convolution structure enables the model to capture the local relationships between adjacent characters and mitigate the influence of pre-trained language models on the text domain. Different-length image convolution kernels are applied in the row and column directions of the grid model to model the Chinese character relationships within evaluation elements.

(3) Experimental results on a Chinese financial commentary text dataset from Jiang et al. \cite{bib6} show that the F1 value of GCGTS is superior to that of the state-of-the-art models such as GTS \cite{bib1} and SDRN \cite{bib2}.

\section{Related Work}\label{sec2}
The concept of sentiment Aspect-Opinion Pair was proposed by Bloom et al. \cite{bib7} in 2007. They defined it as $\langle \text{aspect term, opinion term, evaluation source} \rangle$ and created a rule-based extraction method. This method looks up opinion term  in an existing dictionary and then extracts their corresponding aspect term through rules. While this method is effective to a certain extent, it relies solely on manually constructed rules leading to low efficiency. In the same year, Pang et al. \cite{bib8} proposed the task of Aspect-oriented Fine-grained Opinion Extraction (AFOE) for extracting $\langle \text{aspect term, opinion term } \rangle$, where aspect term represents the phrase in the sentence  referring to a product, service, or entity, and opinion term represents the attitude towards the aspect term in the sentence.

Early sentiment unit extraction was based on rule-based methods that rely on pre-defined rules and external language resources for extraction. Rule-based methods typically observe,  count and analyze data to identify patterns of sentiment Aspect-Opinion Pair in sentences and construct grammar rules. Qiu et al. \cite{bib9} used syntactic dependency rules to enhance the extraction of aspect term and opinion term through double propagation between them. Wu et al. \cite{bib10} suggested that aspect term and opinion term are often represented at the phrase level, rather than the word level, and designed syntax rules based on phrase-level. Zhao et al. \cite{bib11} built a syntax path library and first extracted opinion term using an opinion term dictionary, and then identified sentiment Aspect-Opinion Pair automatically by calling syntax paths. Jiang et al. \cite{bib6} further analyzed dependency syntax relationships and proposed a syntax rule-based sentiment Aspect-Opinion Pair extraction method. Wang et al. \cite{bib12} constructed extraction rules based on part-of-speech, using rules for nouns, verbs, and adjectives to extract evaluation elements, and then combined word collocation relationships to obtain sentiment Aspect-Opinion Pair.

In recent years, more and more scholars have applied deep learning models to the sentiment Aspect-Opinion Pair extraction task. Vo et al. \cite{bib13} proposed using neural networks to learn the hidden word rules in the text, but the network structure used was relatively simple and could not model the relationship between context and opinion term , resulting in poor extraction performance. Sahu et al. \cite{bib6} proposed using BiLSTM for feature extraction and modeling from text, which can capture long-distance context-dependent relationships well, but BiLSTM cannot explicitly locate opinion term or aspect term and does not use syntactic dependency relationships between words. Afterwards, under given a specific aspect term (Target-oriented Opinion Word Extraction, TOWE), Fan et al. \cite{bib15} proposed the Inward-Outward LSTM+Global Context (IOG) model  to extract opinion term, which encodes the target word from both ends of the sentence to the inside using Inward LSTM and from the inside to both ends using Outward LSTM, effectively addressing the problem of BiLSTM's inability to explicitly locate aspect term. Zhu et al. \cite{bib16} modeled the dependency syntax relationships between words in a sentence as a graph and used the graph neural network model GNNs to learn the network. Currently, many research reports at home and abroad use dependency trees \cite{bib17},\cite{bib18},\cite{bib19} to enhance the extraction of Aspect-Opinion Pair, proving that dependency trees  is still worth research in the task of sentiment Aspect-Opinion Pair extraction.

The grid mothod directly models all word pairs in the text as a grid and extracts Aspect-Opinion Pair by judging the category of grid units. In 2020, Wu et al. [1] proposed using a grid-based deep learning model, the Grid Tagging Scheme (GTS), for Aspect-Opinion Pair extraction. Since then, the grid model has been widely used for Aspect-Opinion Pair extraction tasks, for example, Chen et al. \cite{bib2} proposed an enhanced grid model, using a dual-channel approach (opinion term extraction channel and aspect term extraction channel). As the role of syntactic dependencies between Aspect-Opinion Pair is often ignored in early grid models, more and more researchers are focusing on using syntax to enhance grid model extraction performance. Wu et al. \cite{bib4} proposed the PAOTE model, which integrates syntax and part-of-speech information, and considers the influence of part-of-speech on dependency trees. Liang et al. \cite{bib20} argued that previous dependency tree models did not accurately learn the connections between different opinion term in the same sentence, and proposed the BiSyn-GAT+ (Bi-Syntax aware Graph Attention Network) model, which uses constituent trees to model the contextual and sentiment relationships between opinion term and establish relationships between different opinion term. Chen et al. \cite{bib3} believed that current Aspect-Opinion Pair extraction ignores the relationships between words, so they used a multi-channel graph to encode the various dependency relationships between words and proposed the EMC-GCN model. Although this model considers many dependency relationships between words, it ignores the local relationships between adjacent words.

Currently, the grid model is widely used for Aspect-Opinion Pair extraction tasks and achieves good results, but there are still shortcomings in Chinese financial Aspect-Opinion Pair extraction tasks: (1) existing work focuses on English datasets and pays little attention to the characteristics and domain-specific features of Chinese data. (2) When constructing embedding vectors for grid elements, the explicit influence of character-level dependency relationships on grid relationships is ignored. (3) Insufficient learning of Chinese character relationships within evaluation elements.

\section{Graph-based Character-level Grid Tagging Scheme}\label{sec3}

\begin{figure}[h]%
\centering
\includegraphics[width=1.0\textwidth]{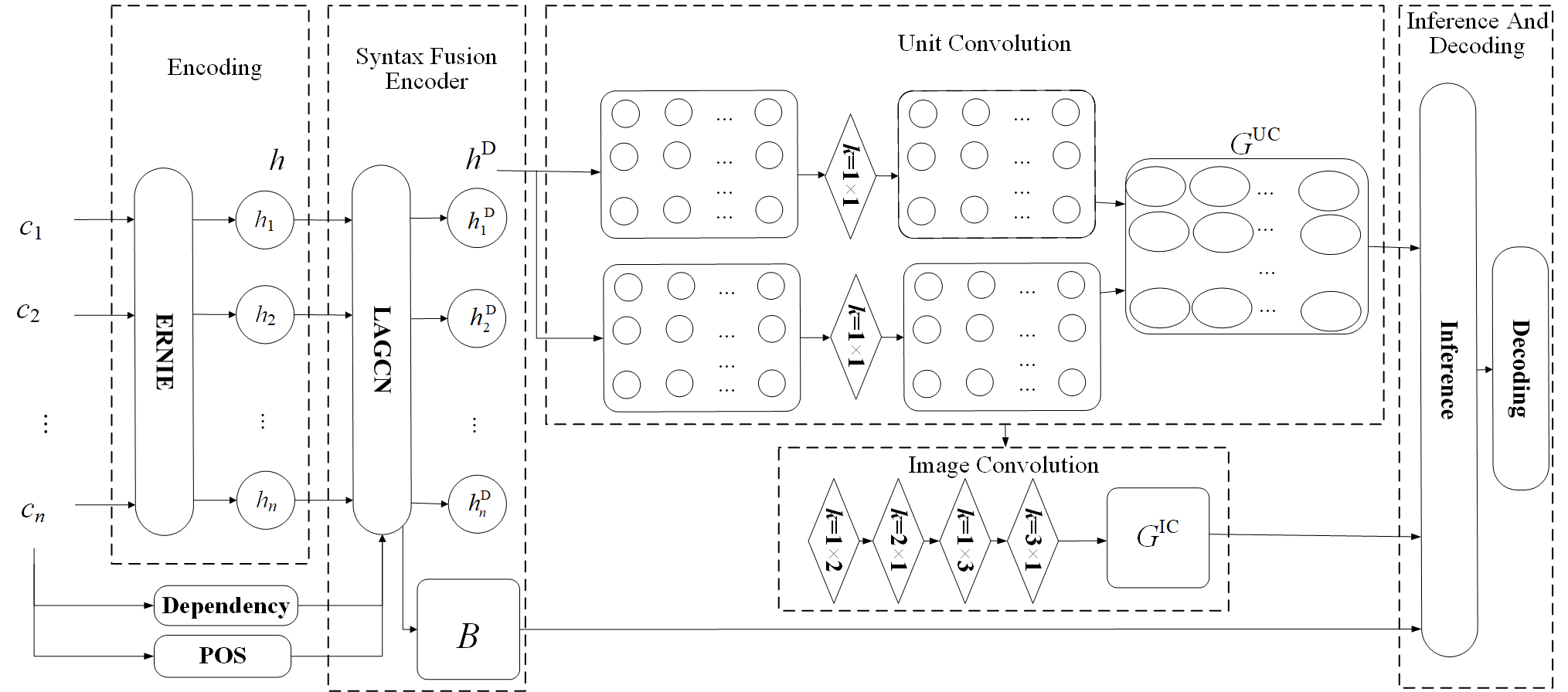}
\caption{The overall architecture of neural models based on Graph-based Character-level Grid Tagging Scheme.}\label{fig2}
\end{figure}

The structure of the Chinese financial Aspect-Opinion Pair extraction model GCGTS based on inter-character relationships is shown in Figure \ref{fig2}, where the diamond block represents convolution operation, and k represents the size of the convolution kernel.

The model uses Baidu's ERNIE\footnotemark[1] encoding for character vectors and HANLP2.0\footnotemark[2] to generate syntactic dependency relationships and POS tags as inputs. First, the character vectors, syntactic dependency relationships, and POS tags are injected into the syntax fusion encoder LAGCN, which outputs a grid tensor B of inter-character relationships and a fused vector $h\textsuperscript{D}$ that incorporates inter-character dependencies (Dependency). Then, two single-unit convolutional kernels are used to differentiate the words in $h\textsuperscript{D}$ transformed into two grids based on rows and columns, so that the same character can have different encoding when it serves as different sentimental evaluation elements. The differentiated tensor $G\textsuperscript{UC}$ is finally output. Here, this encoding method is called Unit Convolution (UC). Next, $G\textsuperscript{UC}$ is convolved in both row and column directions using different-sized image convolutional kernels to output a Chinese character relationship tensor $G\textsuperscript{IC}$ within the sentimental evaluation element. This encoding method is called Image Convolution (IC). Finally, the grid nodes in tensors $B$, $G\textsuperscript{UC}$, and $G\textsuperscript{IC}$ are concatenated, and the decoding layer predicts the labels for each node element.

\footnotetext[1]{https://github.com/PaddlePaddle/ERNIE}
\footnotetext[2]{https://www.hanlp.com/}
\subsection{Encoding}\label{subsec2}

The encoding layer uses ERNIE to encode each character as a separate unit. Given an input sentence $s = \langle c_1, c_2, \ldots, c_n \rangle$, we utilize BERT as the underlying encoder to yield the basic contextualized character representations $h = \langle h_1, h_2, \ldots, h_n \rangle$.

\subsection{Syntax Fusion Encoder}\label{subsec3}
To study inter-character dependency relationships, this article constructs a character adjacency matrix through dependency relationships. The relationship type between two characters is the dependency type between the words they belong to, and the relationship type between characters belonging to the same word is "self." When there is no dependency relationship between the words they belong to, the relationship type between characters is "O".

By aggregating information from other characters that have a dependency relationship with $c_i$, the encoding $h_i$ of character $c_i$ is obtained as follow:
\begin{equation}
h_i^{(l)} = \text{Relu}\left(\sum\limits_{j = 1}^N \alpha_{i,j}^{(l)} (W_1 h_j^{(l-1)}||W_2 r_{i,j} ||W_3 p_j)\right), \tag{1}
\label{eq:equation1}
\end{equation}
The above process is implemented using LAGCN\cite{bib4}. After encoding through L layers of LAGCN, the final result is a fused character vector ${h^{\mathrm{D}}} = [h_1^{\mathrm{D}}, h_2^{\mathrm{D}}, \ldots, h_n^{\mathrm{D}}]$
 and a character relationship grid tensor $B$. The $i$-th row and j-th column of $B$ represent the encoding vector of the dependency relationship between $c_j$ and $c_i$.
The weight of character $c_j$ on character $c_i$ is defined as the inter-word adjacency strength $\alpha_{i,j}$ as shown in follow：
\begin{equation}
\beta _{i,j}^{(l)} = W_4[h_j^{l - 1}||p_j||r_{i,j}], \tag{2}
\label{eq:equation2}
\end{equation}
\begin{equation}
\alpha _{i,j}^{(l)} = \frac{{d_{i,j}\exp \left(\sum(\beta _{i,j}^{(l)})\right)}}{{\sum\limits_{k = 1}^N d_{i,k}\exp \left(\sum(\beta _{i,k}^{(l)})\right)}},
\tag{3}
\label{eq:equation3}
\end{equation}
Where "\(\|\)" denotes the concatenation operation, the encoding of the dependency relationship between $c_i$ and $c_j$, denoted as $\beta _{i,j}$, is obtained from the word vector, part-of-speech, and the dependency relationship of $c_j$, to $c_i$. Where  $\beta _{i,j}^{(l)}$ represents the encoded vector of the $l$-th layer relationship between $c_i$ and $c_j$, $p_j$ represents the encoded POS(Part Of Speech) of character $c_j$ in the word, $r_{i,j}$ represents the encoded dependency relationship category between character $c_i$ and $c_j$. $\beta _{i,j}^\mathrm{D}$ represents the encoding vector for $\beta _{i,j}^{(l)}$ in the final layer of LAGCN, while $\beta _{i,j}^\mathrm{D}$ denotes the entry located at the $i$-th row and $j$-th column of the grid $B$, and $\beta _{i,j}^\mathrm{D}$ represents the encoding vector for $\beta _{i,j}^{(l)}$ in the final layer of LAGCN, while  denotes the embedding located at the $i$-th characters in $h^D$ matrix.

\subsection{Unit Convolution}\label{subsec4}
Previous models differentiated evaluation elements by transposing the grid, which led to incomplete differentiation of evaluation element types. 

\begin{figure}[h]%
\centering
\includegraphics[width=1.0\textwidth]{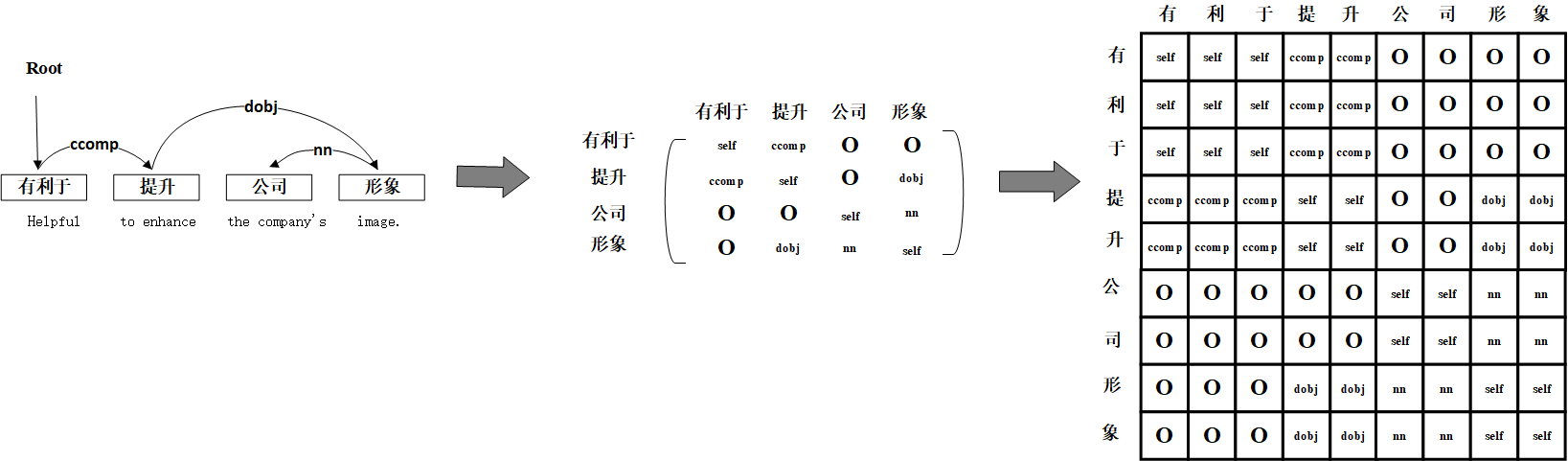}
\caption{ UC operations model.Where the diamond block represents the convolutional operation and $k$ represents the convolutional kernel size.}\label{fig3}
\end{figure}

As shown in Figure \ref{fig3}, in order to further distinguish different types of evaluation elements, this paper adopts two different UC operations on $h^\mathrm{D}$, and concatenates the encoding of the $i$-th row character $c_i$ with the encoding of the $j$-th column character $c_j$ to obtain the element $g_{i,j}^{\mathrm{UC}}$ in the grid $G^{\mathrm{UC}}$ is set as:
\begin{equation}
g_{i,j}^{\mathrm{UC}} = W_{i,j}^{\mathrm{G}}(h_i^{\mathrm{D'}}||h_j^{\mathrm{D''}}).
\tag{4}
\label{eq:equation4}
\end{equation}

\subsection{Image Convolution}\label{subsec5}
In order to learn the relationship between adjacent characters, we use Image Convolution (IC) to obtain representations of candidate evaluation elements. Specifically, we use a 1×$N$ convolutional kernel along the row dimension and a $N$×1 convolution kernel along the column dimension. Considering that words with 1$ \sim $ 3 characters account for 93.94\% of the total after financial text segmentation, we choose $N$=2 and $N$=3 in this paper. Finally, the elements in the grid $G^\mathrm{IC}$ are calculated by performing image convolution on $G^\mathrm{UC}$, where $G^\mathrm{row,2}$ and $G^\mathrm{row,3}$ use 1×2 and 1×3 convolution kernels, and $G^\mathrm{col,2}$ and $G^\mathrm{IC}$ use 2×1 and 3×1 convolution kernels.

\subsection{Inference and Decoding}\label{subsec6}
After fully considering the relationships between words, the paper considers the encoding of the first letter of a word as the final encoding of the word, which is then used to predict labels. This approach not only speeds up the prediction process, but also reduces inconsistencies caused by excessive labeling. The GCGTS model retains GTS's inference capabilities and takes into account the constraints between labels. Through multiple rounds of inference, the distribution of letters to labels in the previous round's global scope affects the inference of letters to labels in the next round.

The predicted probability distribution $p_{i,j}^t$ of characters-pair $\langle \text{$c_i$,$c_j$} \rangle$
 can be calculated as follows:

\begin{equation}
p_{i,j}^0 = g_{i,j}^{\mathrm{IC}},
\tag{5}
\label{eq:equation5}
\end{equation}
\begin{equation}
z_{i,j}^0 = W_1^{\mathrm{I}}(h_i^{D'}||h_j^{D''}||\beta _{i,j}^D),
\tag{6}
\label{eq:equation6}
\end{equation}
\begin{equation}
z_{i,j}^t = W_2^{\mathrm{I}}(z_{i,j}^{t-1}||\mathrm{maxpooling}(p_{i}^{t-1})||\mathrm{maxpooling}(p_{j}^{t-1})||p_{i,j}^{t-1}),
\tag{7}
\label{eq:equation7}
\end{equation}
\begin{equation}
p_{i,j}^t = \mathrm{softmax}(W_3^{\mathrm{I}}z_{i,j}^t + b_2^{\mathrm{I}}).
\tag{8}
\label{eq:equation8}
\end{equation}
In the above process, the element in the $i$-th row and $j$-th column of $G^\mathrm{IC}$ is denoted as $g_{i,j}^{{\rm{IC}}}$.Where t represents the iteration round, $p_{i,j}^t$ denotes the predicted probability distribution of the word pair relation $\langle \text{$c_i$,$c_j$} \rangle$ in the t-th round, and $z_{i,j}^t$ represents the feature representation of the word pair relation $\langle \text{$c_i$,$c_j$} \rangle$ in the $t$-th round.

\subsection{Training Loss}\label{subsec7}
The cross-entropy loss of GCGTS is as follows:
\begin{equation}
\text{loss} = - \sum\limits_{i=1}^{n} \sum\limits_{j=i}^{n} \sum\limits_{k \in C} \mathbb{I}({y_{i,j}} = q) \log(p_{i,j|q}^T),
\tag{9}
\label{eq:equation9}
\end{equation}
where I($y_{i,j}$=$q$) is an indicator function, $C$ is {A, O, P, E}, $p_{i,j}^T$ represents the predicted probability distribution of the last round of word-pair relationships $\langle \text{$c_i$,$c_j$} \rangle$ with the probability of label $q$, and $y_{i,j}$ represents the true label of the word-pair relationship $\langle \text{$c_i$,$c_j$} \rangle$. 

\section{Experiments}\label{sec4}
\subsection{Datasets}\label{subsec8}
The dataset used in this article is taken from the manufacturing sector of the financial review dataset constructed by Jiang et al. \cite{bib6}. The dataset was manually corrected and divided into training, validation, and testing sets with an 8:1:1 ratio. The evaluation elements and Aspect-Opinion Pair of each dataset are summarized in Table \ref{table1}.

\begin{table}[h]
\caption{Statistics of evaluation elements and Aspect-Opinion Pair information}
\label{table1}
\begin{tabular}{ccccc}
\toprule
\textbf{Dataset} &\textbf{Sentence } & \textbf{Aspect Term} & \textbf{Opinion Term} & \textbf{Aspect-Opinion Pair } \\
\midrule
train & 2293 & 3473 & 3411 & 3606 \\
dev & 322 & 473& 464  & 490 \\
test & 277 & 422 & 410 & 436 \\
total & 2892 & 4368 & 3867 & 4532 \\

\botrule
\end{tabular}

\end{table}

\subsection{Compared Methods}\label{subsec9}

This chapter uses the Pytorch framework under the Linux environment to build the model. 
At inference time,  we set  parametersthe Bach\_size =12, label\_category=4 ,lr=5e-5 and the number of layers in LAGCN is set to 2.

The following section provides an exposition of the reference method adopted in this article：

(1) BiLSTM model (Bidirectional Long Short-Term Memory) \cite{bib21}. This model is one of the most commonly used models in named entity recognition tasks. It uses bidirectional LSTMs to learn contextual features for label classification. In this paper, we use this model to extract aspect term.

(2) BiLSTM-Att-CRF (Bidirectional Long Short-Term Memory Attention Conditional Random Field) \cite{bib22}.  By integrating BiLSTM with attention mechanism and conditional random field, the model can learn more features.

(3) IOG model \cite{bib15}. This model is one of the representative models of the Pipline Model method. The model aims to extract opinion term by evaluating objects and is regarded as a typical model  to extract Aspect-Opinion Pair by some works \cite{bib1},\cite{bib3}.

(4) SDRN model \cite{bib2}. This model extracts aspect term, opinion term, and Aspect-Opinion Pair through multi-task learning, using a grid model and labelling the Aspect-Opinion Pair on each token, which yields good results.

(5) GTS model \cite{bib1}. Using a grid model, it is one of the representative baseline Models and considered as an advanced model for extracting Aspect-Opinion Pair in related works \cite{bib3},\cite{bib4}.

(6) GTS-Char model. Unlike GTS, which only uses the first character of the word for decoding, GTS-Char uses all the characters in the word for decoding, which is similar to the SDRN model.

\subsection{Results of Aspect-Opinion Pair Extraction}\label{subsec10}
The evaluation of various models on the Chinese financial review text dataset was conducted using Baidu ERNIE as the pre-trained model, and the results are shown in Table \ref{table2}. Since error propagation existed, the effectiveness of the pipeline model in extracting evaluation targets or words theoretically is not better than that of separate extraction. Therefore, only the effectiveness of the pipeline model in extracting Aspect-Opinion Pair is listed in the table.

\begin{table}[h]
\centering
\caption{Results of comparison experimental (unit:\%)}
\label{table2}
\setlength{\tabcolsep}{1.5mm}{
\begin{tabular}{cccccccccc}
\hline
\textbf{Methods} & \multicolumn{3}{c}{Aspect-Opinion Pair} & \multicolumn{3}{c}{Aspect Term} & \multicolumn{3}{c}{Opinion Term} \\ \cmidrule(lr){2-4} \cmidrule(lr){5-7} \cmidrule(lr){8-10}
& P & R & F1 & P & R & F1 & P & R & F1 \\ 
\midrule

BiLSTM & - & - & - & 69.10 & 60.91 & 64.75 & - & - & - \\
BiLSTM-CRF & - & - & - & 71.80 & 60.23 & 65.51 & - & - & - \\
BiLSTM-Att-CRF & - & - & - & 71.88 & 60.91 & 65.94 & - & - & - \\
IOG & - & - & - & - & - & - & 72.65 & 73.47 & 73.06 \\
\midrule
\multicolumn{10}{c}{ Pipline Models} \\
\midrule
BiLSTM+IOG & 42.35 & 41.28 & 41.81 & - & - & - & - & - & - \\
BiLSTM-CRF+IOG & 45.45 & 41.28 & 43.27 & - & - & - & - & - & - \\
BiLSTM-Att-CRF+IOG & 45.27 & 41.74 & 43.44 & - & - & - & - & - & - \\
\midrule
\multicolumn{10}{c}{ Joint Models} \\
\midrule
SDRN & 59.51 & 61.01 & 60.25 & 67.43 & 69.53 & 68.52 & 77.06 & 84.60 & 80.65 \\
GTS-Char & 61.79 & 62.16 & 61.94 & 66.16 & 72.66 & 69.26 & 79.17 & \textbf{88.70} & \textbf{83.66} \\
GTS & 60.70 & 63.72 & 62.15 & 66.08 & 72.42 & 69.11 & 79.09 & 85.50 & 82.17 \\
GCGTS & \textbf{62.82} & \textbf{64.04} & \textbf{63.41} & \textbf{70.75} & \textbf{74.82} & \textbf{72.73} & \textbf{80.73} & 86.49 & 83.51 \\
\botrule
\end{tabular}}
\end{table}

It is observed that the GCGTS model outperformed all the compared models on the Chinese financial text dataset, achieving a 2.03\% improvement over the classic GTS model and a 5.24\% improvement over the grid model SDRN. This indicates that the proposed method of enhancing the inter-character relationships is effective. The joint learning model showed significant improvements in evaluating unit extraction, evaluating word extraction, and evaluating object extraction, compared to the pipeline model, due to its error propagation during the step-by-step extraction process. 

In both the GTS-Char and SDRN models, all the characters of the sentence participated in decoding, but the GTS-Char model showed better performance in F1 score than SDRN, because the GTS-Char model had an additional inference stage. The GTS model had an advantage in evaluating unit extraction, because it required fewer predicted labels compared to the GTS-Char model, in which only the first character of each word participate in the annotation decoding. It is worth noting that although the GTS-Char model did not show advantages in evaluating unit extraction, it demonstrated good performance in evaluating object and word extraction tasks.

\subsection{Ablation Study}\label{subsec11}
In order to verify the effects of each module in the GCGTS model, this section conducted ablation experiments on the model. Taking the GTS model as the baseline, we added the Unit Convolution (UC), Image Convolution (IC), and Syntax Fusion Encoder LAGCN modules separately. The Syntax Fusion Encoder LAGCN was divided into two steps: using the word vectors $h^D$ (DGTS, Dependency Grid Tagging Scheme)  with the inter-character dependencies alone, and with both inter-character dependencies and the grid tensor $B$ (DBGTS, Dependency Binding-char Grid Tagging Scheme) representing inter-character relationships. The results of the Ablation experiment as shown in Table \ref{table3}.

\begin{table}[h]
  \caption{Results of the Ablation experiment (unit: \%)}
  \label{table3}
  \setlength{\tabcolsep}{3mm}{
    \begin{tabular}{cccc}
    \toprule
          & P     & R     & F1 \\
    \midrule
    GTS   & 60.70 & 63.72 & 62.15 \\
    GTS+UC & 62.27 & 63.56 & 62.89 (+1.19) \\
    GTS+IC & 62.11 & \textbf{64.29} & 63.15 (+1.64) \\
    DGTS  & 61.38 & 63.90 & 62.60 (+0.72) \\
    DBGTS & 61.63 & 64.28 & 62.92 (+1.24) \\
    GCGTS & \textbf{62.82} & 64.04 & \textbf{63.41} (+2.03) \\
    \bottomrule
    \end{tabular}}
\end{table}

After incorporating the corresponding modules, notable improvements in the performance of the GCGTS model were observed. When UC was integrated into GTS, the F1 score increased by 1.19\%, indicating that the approach used to distinguish different types of evaluation elements had a positive effect. Similarly, the inclusion of the inter-character relationship modeling module IC in GTS resulted in a 1.64\% increase in the F1 score, providing supplementary information for identifying sentiment Aspect-Opinion Pairs within evaluation elements.

The introduction of the GCN module to encode the dependency syntax of evaluation elements (DGTS, Dependency-based Graph Tagging Scheme) enabled GTS to fully consider the relationships between these elements, leading to a 0.72\% improvement in the F1 score. Furthermore, by directly incorporating the inter-character relation grid tensor B (DBGTS, Dependency-based Boundary-aware Graph Tagging Scheme) into GTS, the F1 score continued to improve, reaching 1.24\%. This demonstrates that the adjacency matrix effectively aids the model in identifying evaluation object boundaries and further enhances the extraction of Aspect-Opinion Pairs.

In summary, compared to the GTS model, the GCGTS model demonstrated an overall improvement of 2.03\% in performance. These results highlight the effectiveness of incorporating the UC, IC, GCN, and DBGTS modules, which collectively address the shortcomings of the GTS model and enhance the extraction of Aspect-Opinion Pairs from Chinese financial texts.

\subsection{Case study}\label{subsec12}

To validate the effectiveness of the GCGTS model, this study conducted an analysis of extraction examples using a Chinese financial review text dataset, as presented in Table \ref{table4}.

In Example 1, the GTS model failed to accurately extract both the opinion term and the aspect term due to the complexity of the sentence. In contrast, the GCGTS model successfully captured the complete aspect term, indicating its advantage in handling complex sentence structures.

In Example 2, both the GTS and GCGTS models were able to extract the opinion term, but only the GCGTS model managed to retrieve the complete aspect term. This observation demonstrates the effectiveness of the GCGTS model in learning Chinese character relationships within evaluation elements.

In Example 3, both models achieved satisfactory results as the aspect term and opinion term were relatively simple, further validating their capabilities in handling straightforward cases.

However, in Example 4, the extraction results of both models were unsatisfactory due to the length of the aspect term, suggesting the need for further improvements in the GCGTS model to enhance the extraction of longer aspect terms.

Based on the analysis of these extraction examples, we can conclude that the GCGTS model exhibits clear advantages over the traditional GTS model in the extraction of evaluation elements (i.e., aspect terms and opinion terms) from Chinese financial review texts. The GCGTS model demonstrates the ability to adapt to complex sentence structures and effectively learn and utilize Chinese character relationships within evaluation elements. These findings are of significant academic and practical importance in the domains of Chinese financial information processing and sentiment analysis, providing valuable insights for the future refinement and optimization of the GCGTS model.

\begin{table}[h]
  \centering
  \caption{Example of GCGTS model extraction (The aspect term is in bold and the opinion term is underlined)}
  \label{table4}
    \begin{tabular}{l p{4cm} p{3cm} p{3cm}}
    \toprule
    Examples & Sentences & GTS & GCGTS \\
    \midrule
    1 & 公司加强了对\textbf{管理费用}的控制，使其同比\underline{减少}1.52\%。 & 未识别 & 管理费用，减少 \\
    & (The company strengthened its control over \textbf{management expenses}, resulting in a year-on-year \underline{decrease} of 1.52\%.)&unrecognizable &  management expenses, decrease \\
    \midrule
    2 & 公司\underline{扩张}了\textbf{营销队伍}。 & 营销，扩张 & 营销队伍，扩张 \\
    & (The company \underline{expanded} its \textbf{marketing team}.) & marketing, expanded & marketing team, expanded \\
    \midrule
    3 & 公司2010年实现\textbf{营业收入}1.46亿元，同比\underline{下降}14.62\%。 & 营业收入，下降 & 营业收入，下降 \\
    & (The company achieved \textbf{a total operating revenue} of 146 million yuan in 2010, representing a year-on-year \underline{decrease} of 14.62\%.) & a total operating revenue, decrease & a total operating revenue, decrease \\
    \midrule
    4 & \textbf{户外鞋品和户外装备收入增速}分别为49.6\%和42.3\%，\underline{低于}户外服装的收入增速。 & 收入，低于 & 收入，低于 \\
    & (\textbf{The revenue growth rates for outdoor footwear and outdoor equipment} were 49.6\% and 42.3\%, respectively, both \underline{lower} than the revenue growth rate for outdoor clothing.) & revenue, lower &revenue, lower \\
    \bottomrule
    \end{tabular}%
  \label{tab:example}%
\end{table}%

\begin{figure}[h]%
\centering
\includegraphics[width=0.5\textwidth]{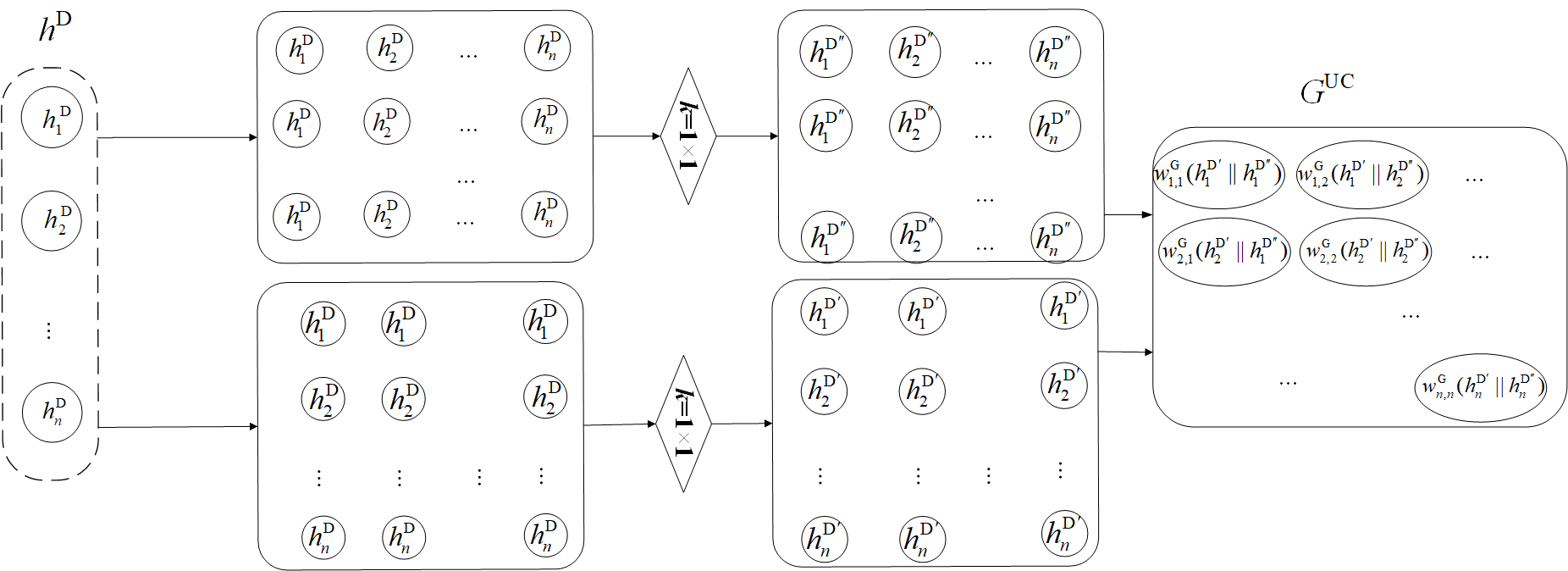}
\caption{ Grayscale image of adjacency tensor in GCGTS model.It is a visualization of the tensor matrix of the model's final predicted Aspect-Opinion Pair. It can reflect the training status of each node and judge model performance.The horizontal axis represents the opinion term, and the vertical axis represents the aspect term. The darker  of the grid color, the greater the probability that it pairs aspect and opinion.}\label{fig4}
\end{figure}

\subsection{Adjacency Tensor Graph Study}\label{subsec13}
The adjacency tensor graph is a visualization of the tensor matrix of the model's final predicted Aspect-Opinion Pair. It can reflect the training status of each node and judge model performance. Figure \ref{fig4} shows the adjacency tensor graph obtained by GCGTS model on sentence "偿债 能力 、 营运 能力 较强。''(Debt repayment ability and operating capability are strong.) The Aspect-Opinion Pair in this sentence include ``$\langle \text{偿债 能力，较强} \rangle$''($\langle \text{debt repayment ability, strong} \rangle$) and ``$\langle \text{营运 能力，较强} \rangle$''($\langle \text{operating capability, strong} \rangle$).

In Figure \ref{fig4}, since this paper predicts Aspect-Opinion Pair using the first tag of word, the color of the first letter in the Aspect-Opinion Pair is darker. For instance, it can be observed that the grids at the ``较强''(strong) column, and at the``偿债 能力''(debt repayment ability) and the ``营运 能力''(operating capability) rows are darker, indicating that the GCGTS model can better distinguish these pairs of Aspect-Opinion Pair.

\section{Conclusions and Future Work}\label{sec5}
In recent years, the introduction of grid models and syntactic dependency relationships has become a research trend in  evaluating unit extraction. This article focuses on the extraction task from Chinese financial texts. It first introduces LAGCN to learn long-distance syntactic dependencies in Chinese financial texts. Using adjacency matrix to enhance the problem of fuzzy boundaries in evaluation elements extraction.. The article then proposes the use of Unit Convolution (UC) to differentiate the aspect term and opinion term, and  the use of Image Convolution (IC) to fully learn the relationships between adjacent words within words and evaluation elements, while the latter could further differentiating the aspect term and opinion term.

Future research can improve and expand the GCGTS model in several aspects. Firstly, it is necessary to address the challenge of handling long evaluation targets in the grid model. Possible approaches include designing more flexible network architectures, optimizing input representation methods, or introducing more complex attention mechanisms. Secondly, as this study only considered the disorderliness of financial domain terms and the complexity of syntactic unit relationships in financial texts, it did not leverage additional external financial knowledge and contextual information for extracting financial evaluation elements. Future research can explore effective ways to utilize domain knowledge, dictionaries, knowledge graphs, and incorporate contextual information into the model to enhance the performance and accuracy of the extraction task. In conclusion, by improving the model's structure, considering diverse application requirements in different domains and languages, and leveraging external knowledge and contextual information, further advancements can be achieved in the research and application of aspect opinion extraction in Chinese financial texts, providing stronger support for information processing and decision-making in the financial domain.



\bibliography{sn-bibliography}


\begin{thebibliography}{21}
\ifx \bisbn   \undefined \def \bisbn  #1{ISBN #1}\fi
\ifx \binits  \undefined \def \binits#1{#1}\fi
\ifx \bauthor  \undefined \def \bauthor#1{#1}\fi
\ifx \batitle  \undefined \def \batitle#1{#1}\fi
\ifx \bjtitle  \undefined \def \bjtitle#1{#1}\fi
\ifx \bvolume  \undefined \def \bvolume#1{\textbf{#1}}\fi
\ifx \byear  \undefined \def \byear#1{#1}\fi
\ifx \bissue  \undefined \def \bissue#1{#1}\fi
\ifx \bfpage  \undefined \def \bfpage#1{#1}\fi
\ifx \blpage  \undefined \def \blpage #1{#1}\fi
\ifx \burl  \undefined \def \burl#1{\textsf{#1}}\fi
\ifx \doiurl  \undefined \def \doiurl#1{\url{https://doi.org/#1}}\fi
\ifx \betal  \undefined \def \betal{\textit{et al.}}\fi
\ifx \binstitute  \undefined \def \binstitute#1{#1}\fi
\ifx \binstitutionaled  \undefined \def \binstitutionaled#1{#1}\fi
\ifx \bctitle  \undefined \def \bctitle#1{#1}\fi
\ifx \beditor  \undefined \def \beditor#1{#1}\fi
\ifx \bpublisher  \undefined \def \bpublisher#1{#1}\fi
\ifx \bbtitle  \undefined \def \bbtitle#1{#1}\fi
\ifx \bedition  \undefined \def \bedition#1{#1}\fi
\ifx \bseriesno  \undefined \def \bseriesno#1{#1}\fi
\ifx \blocation  \undefined \def \blocation#1{#1}\fi
\ifx \bsertitle  \undefined \def \bsertitle#1{#1}\fi
\ifx \bsnm \undefined \def \bsnm#1{#1}\fi
\ifx \bsuffix \undefined \def \bsuffix#1{#1}\fi
\ifx \bparticle \undefined \def \bparticle#1{#1}\fi
\ifx \barticle \undefined \def \barticle#1{#1}\fi
\bibcommenthead
\ifx \bconfdate \undefined \def \bconfdate #1{#1}\fi
\ifx \botherref \undefined \def \botherref #1{#1}\fi
\ifx \url \undefined \def \url#1{\textsf{#1}}\fi
\ifx \bchapter \undefined \def \bchapter#1{#1}\fi
\ifx \bbook \undefined \def \bbook#1{#1}\fi
\ifx \bcomment \undefined \def \bcomment#1{#1}\fi
\ifx \oauthor \undefined \def \oauthor#1{#1}\fi
\ifx \citeauthoryear \undefined \def \citeauthoryear#1{#1}\fi
\ifx \endbibitem  \undefined \def \endbibitem {}\fi
\ifx \bconflocation  \undefined \def \bconflocation#1{#1}\fi
\ifx \arxivurl  \undefined \def \arxivurl#1{\textsf{#1}}\fi
\csname PreBibitemsHook\endcsname

\bibitem[\protect\citeauthoryear{Wu et~al.}{2020}]{bib1}
\begin{botherref}
\oauthor{\bsnm{Wu}, \binits{Z.}},
\oauthor{\bsnm{Ying}, \binits{C.}},
\oauthor{\bsnm{Zhao}, \binits{F.}}, et al.:
Grid Tagging Scheme for Aspect-oriented Fine Grained.
Proceedings of the 2020 Conference on Empirical Methods in Natural Language
  Processing, Online, 2576--2585 November 2020
(2020)
\end{botherref}
\endbibitem

\bibitem[\protect\citeauthoryear{Chen et~al.}{2020}]{bib2}
\begin{botherref}
\oauthor{\bsnm{Chen}, \binits{S.}},
\oauthor{\bsnm{Liu}, \binits{J.}},
\oauthor{\bsnm{Wang}, \binits{Y.}}, et al.:
Synchronous Double-channel Recurrent Network for Aspect-Opinion Pair
  Extraction.
Proceedings of the 58th Annual Meeting of the Association for Computational
  Linguistics, Online, 6515--6524 July 2020
(2020)
\end{botherref}
\endbibitem

\bibitem[\protect\citeauthoryear{Chen et~al.}{2022}]{bib3}
\begin{botherref}
\oauthor{\bsnm{Chen}, \binits{H.}},
\oauthor{\bsnm{Zhai}, \binits{Z.}},
\oauthor{\bsnm{Feng}, \binits{F.}}, et al.:
Enhanced Multi-Channel Graph Convolutional Network for Aspect Sentiment Triplet
  Extraction.
Proceedings of the 60th Annual Meeting of the Association for Computational
  Linguistics, Dublin, Ireland, 2974--2985 May 2022
(2022)
\end{botherref}
\endbibitem

\bibitem[\protect\citeauthoryear{Wu et~al.}{2021}]{bib4}
\begin{botherref}
\oauthor{\bsnm{Wu}, \binits{S.}},
\oauthor{\bsnm{Fei}, \binits{H.}},
\oauthor{\bsnm{Ren}, \binits{Y.}}, et al.:
Learn from Syntax: Improving Pair-wise Aspect and Opinion Terms Extraction with
  Rich Syntactic Knowledge.
Proceedings of the 30th International Joint Conference on Artificial
  Intelligence, Québec, Canada, 21 August 2021
(2021)
\end{botherref}
\endbibitem

\bibitem[\protect\citeauthoryear{Zhang et~al.}{2019}]{bib5}
\begin{botherref}
\oauthor{\bsnm{Zhang}, \binits{C.}},
\oauthor{\bsnm{Li}, \binits{Q.}},
\oauthor{\bsnm{Song}, \binits{D.}}:
Aspect-based Sentiment Classification with Aspect-specific Graph Convolutional
  Networks.
Proceedings of the 2019 Conference on Empirical Methods in Natural Language
  Processing and the 9th International Joint Conference on Natural Language
  Processing, Hong Kong, China, 4568--4578 November 2019
(2019)
\end{botherref}
\endbibitem

\bibitem[\protect\citeauthoryear{Jiang et~al.}{2017}]{bib6}
\begin{barticle}
\bauthor{\bsnm{Jiang}, \binits{T.J.}},
\bauthor{\bsnm{Wan}, \binits{C.X.}},
\bauthor{\bsnm{Liu}, \binits{D.X.}}, \betal:
\batitle{Extracting target-opinion pairs based on semantic analysis}.
\bjtitle{Chinese Journal of Computers}
\bvolume{40}(\bissue{3}),
\bfpage{617}--\blpage{633}
(\byear{2017})
\end{barticle}
\endbibitem

\bibitem[\protect\citeauthoryear{Bloom et~al.}{2007}]{bib7}
\begin{botherref}
\oauthor{\bsnm{Bloom}, \binits{K.}},
\oauthor{\bsnm{Garg}, \binits{N.}},
\oauthor{\bsnm{Argamon}, \binits{S.}}:
Extracting Appraisal Expressions.
Proceedings of the Human Language Technology Conference and the Conference on
  Empirical Methods in Natural Language Processing, British Columbia, Canada ,
  308--315 October 2005
(2007)
\end{botherref}
\endbibitem

\bibitem[\protect\citeauthoryear{Pang and Lee}{2007}]{bib8}
\begin{barticle}
\bauthor{\bsnm{Pang}, \binits{B.}},
\bauthor{\bsnm{Lee}, \binits{L.}}:
\batitle{Opinion mining and sentiment analysis}.
\bjtitle{Foundations and Trends in Information Retrieval}
\bvolume{2}(\bissue{1-2}),
\bfpage{1}--\blpage{135}
(\byear{2007})
\end{barticle}
\endbibitem

\bibitem[\protect\citeauthoryear{Qiu et~al.}{2011}]{bib9}
\begin{barticle}
\bauthor{\bsnm{Qiu}, \binits{G.}},
\bauthor{\bsnm{Liu}, \binits{B.}},
\bauthor{\bsnm{Bu}, \binits{J.}}, \betal:
\batitle{Opinion word expansion and target extraction through double
  propagation}.
\bjtitle{Computational Linguistics}
\bvolume{37}(\bissue{1}),
\bfpage{9}--\blpage{27}
(\byear{2011})
\end{barticle}
\endbibitem

\bibitem[\protect\citeauthoryear{Wu et~al.}{2009}]{bib10}
\begin{botherref}
\oauthor{\bsnm{Wu}, \binits{Y.}},
\oauthor{\bsnm{Zhang}, \binits{Q.}},
\oauthor{\bsnm{Huang}, \binits{X.J.}}, et al.:
Phrase Dependency Parsing for Opinion Mining.
Proceedings of the 2009 Conference on Empirical Methods in Natural Language
  Processing, Singapore, 1533--1541 August 2009
(2009)
\end{botherref}
\endbibitem

\bibitem[\protect\citeauthoryear{Zhao et~al.}{2011}]{bib11}
\begin{barticle}
\bauthor{\bsnm{Zhao}, \binits{Y.Y.}},
\bauthor{\bsnm{Qin}, \binits{B.}},
\bauthor{\bsnm{Che}, \binits{W.X.}}, \betal:
\batitle{Appraisal expression recognition based on syntactic path}.
\bjtitle{Journal of Software}
\bvolume{22}(\bissue{5}),
\bfpage{887}--\blpage{898}
(\byear{2011})
\end{barticle}
\endbibitem

\bibitem[\protect\citeauthoryear{Wang and Wu}{2012}]{bib12}
\begin{barticle}
\bauthor{\bsnm{Wang}, \binits{S.G.}},
\bauthor{\bsnm{Wu}, \binits{S.H.}}:
\batitle{Feature-opinion extraction in scenic spots reviews based on dependency
  relation}.
\bjtitle{Journal of Chinese Information Processing}
\bvolume{26}(\bissue{3}),
\bfpage{116}--\blpage{121}
(\byear{2012})
\end{barticle}
\endbibitem

\bibitem[\protect\citeauthoryear{Vo and Zhang}{2015}]{bib13}
\begin{botherref}
\oauthor{\bsnm{Vo}, \binits{D.T.}},
\oauthor{\bsnm{Zhang}, \binits{Y.}}:
Target-dependent Twitter Sentiment Classification with Rich Automatic Features.
Proceedings of the 24th International Conference on Artificial Intelligence,
  Buenos Aires, Buenos Aires, Argentina, 1347-1353 July 2015
(2015)
\end{botherref}
\endbibitem

\bibitem[\protect\citeauthoryear{Fan et~al.}{2019}]{bib15}
\begin{botherref}
\oauthor{\bsnm{Fan}, \binits{Z.}},
\oauthor{\bsnm{Wu}, \binits{Z.}},
\oauthor{\bsnm{Dai}, \binits{X.Y.}}, et al.:
Target-oriented Opinion Words Extraction with Target-fused Neural Sequence
  Labeling.
Proceedings of the 2019 Conference of the North American Chapter of the
  Association for Computational Linguistics, Minneapolis, America, 2509--2518
  June 2019
(2019)
\end{botherref}
\endbibitem

\bibitem[\protect\citeauthoryear{Zhu et~al.}{2019}]{bib16}
\begin{botherref}
\oauthor{\bsnm{Zhu}, \binits{H.}},
\oauthor{\bsnm{Lin}, \binits{Y.}},
\oauthor{\bsnm{Liu}, \binits{Z.}}, et al.:
Graph Neural Networks with Generated Parameters for Relation Extraction.
Proceedings of the 57th Annual Meeting of the Association for Computational
  Linguistics,Florence, Italy, 1331--1339 July 2019
(2019)
\end{botherref}
\endbibitem

\bibitem[\protect\citeauthoryear{Zhang et~al.}{2019}]{bib17}
\begin{botherref}
\oauthor{\bsnm{Zhang}, \binits{C.}},
\oauthor{\bsnm{Li}, \binits{Q.}},
\oauthor{\bsnm{Song}, \binits{D.}}:
Aspect-based Sentiment Slassification with Aspect-specific Graph Convolutional
  Networks.
Proceedings of the 2019 Conference on Empirical Methods in Natural Language
  Processing and the 9th International Joint Conference on Natural Language
  Processing, Hong Kong, China, 4567--4577 November 2019
(2019)
\end{botherref}
\endbibitem

\bibitem[\protect\citeauthoryear{Wang et~al.}{2020}]{bib18}
\begin{botherref}
\oauthor{\bsnm{Wang}, \binits{K.}},
\oauthor{\bsnm{Shen}, \binits{W.}},
\oauthor{\bsnm{Yang}, \binits{Y.}}, et al.:
Relational Graph Attention Network for Aspect-based Sentiment Analysis.
Proceedings of the 58th Annual Meeting of the Association for Computational
  Linguistics, Online, 3229--3238 July 2020
(2020)
\end{botherref}
\endbibitem

\bibitem[\protect\citeauthoryear{Chen et~al.}{2022}]{bib19}
\begin{botherref}
\oauthor{\bsnm{Chen}, \binits{C.}},
\oauthor{\bsnm{Teng}, \binits{Z.}},
\oauthor{\bsnm{Wang}, \binits{Z.}}, et al.:
Discrete Opinion Tree Induction for Aspect-based Sentiment Analysis.
Proceedings of the 60th Annual Meeting of the Association for Computational
  Linguistics,Dublin, Ireland, 2051--2064 May 2022
(2022)
\end{botherref}
\endbibitem

\bibitem[\protect\citeauthoryear{Huang et~al.}{2020}]{bib20}
\begin{botherref}
\oauthor{\bsnm{Huang}, \binits{L.}},
\oauthor{\bsnm{Sun}, \binits{X.}},
\oauthor{\bsnm{Li}, \binits{S.}}, et al.:
Syntax-aware Graph Attention Network for Aspect-level Sentiment Classification.
Proceedings of the 28th International Conference on Computational Linguistics,
  Barcelona, Spain , 799--810 December 2020
(2020)
\end{botherref}
\endbibitem

\bibitem[\protect\citeauthoryear{Graves et~al.}{2013}]{bib21}
\begin{bchapter}
\bauthor{\bsnm{Graves}, \binits{A.}},
\bauthor{\bsnm{Jaitly}, \binits{N.}},
\bauthor{\bsnm{Mohamed}, \binits{A.}}:
\bctitle{Hybrid speech recognition with deep bidirectional lstm}.
In: \bbtitle{2013 IEEE Workshop on Automatic Speech Recognition and
  Understanding},
pp. \bfpage{273}--\blpage{278}
(\byear{2013})
\end{bchapter}
\endbibitem

\bibitem[\protect\citeauthoryear{Lample et~al.}{2016}]{bib22}
\begin{bchapter}
\bauthor{\bsnm{Lample}, \binits{G.}},
\bauthor{\bsnm{Ballesteros}, \binits{M.}},
\bauthor{\bsnm{Subramanian}, \binits{S.}}, \betal:
\bctitle{Neural architectures for named entity recognition}.
In: \bbtitle{Proceedings of the 14th North American Chapter of the Association
  for Computational Linguistics: Human Language Technologies},
pp. \bfpage{260}--\blpage{270}
(\byear{2016})
\end{bchapter}
\endbibitem

\end{thebibliography}

\end{CJK*}
\end{document}